\documentclass[10pt,twocolumn,letterpaper]{article}

\usepackage{iccv}
\usepackage{times}
\usepackage{epsfig}
\usepackage{graphicx}
\usepackage{amsmath}
\usepackage{amssymb}

\usepackage{amsmath,amsfonts,bm}

\def\eqref#1{equation~\ref{#1}}

\def\1{\bm{1}}

\DeclareMathAlphabet{\mathsfit}{\encodingdefault}{\sfdefault}{m}{sl}
\SetMathAlphabet{\mathsfit}{bold}{\encodingdefault}{\sfdefault}{bx}{n}

\usepackage{xfrac}

\usepackage{url}
\usepackage{enumitem}
\usepackage{wrapfig}
\usepackage{booktabs}       
\usepackage{graphicx}
\usepackage{subcaption}     
\usepackage{algorithm}
\usepackage{algpseudocode}
\usepackage{multirow}
\usepackage{xcolor}
\usepackage{amsthm}
\usepackage{amsmath}
\usepackage{marginnote}

\usepackage[pagebackref=true,breaklinks=true,letterpaper=true,colorlinks,bookmarks=false]{hyperref}

\iccvfinalcopy 

\ificcvfinal\fi

\begin{document}

\title{StackMix: A complementary Mix algorithm}

\author{John Chen\\
Rice University\\
{\tt\small johnchen@rice.edu}
\and
Samarth Sinha\\
University of Toronto\\
{\tt\small samarth.sinha@mail.utoronto.ca}
\and
Anastasios Kyrillidis\\
Rice University\\
{\tt\small anastasios@rice.edu}
}

\maketitle
\ificcvfinal\thispagestyle{empty}\fi

\begin{abstract}
Techniques combining multiple images as input/output have proven to be effective data augmentations for training convolutional neural networks. 
In this paper, we present StackMix: Each input is presented as a concatenation of two images, and the label is the mean of the two one-hot labels. 
On its own, StackMix rivals other widely used methods in the ``Mix'' line of work. 
More importantly, unlike previous work, significant gains across a variety of benchmarks are achieved by combining StackMix with existing Mix augmentation, effectively mixing more than two images.
E.g., by combining StackMix with CutMix, test error in the supervised setting is improved across a variety of settings over CutMix, including 0.8\% on ImageNet, 3\% on Tiny ImageNet, 2\% on CIFAR-100, 0.5\% on CIFAR-10, and 1.5\% on STL-10. Similar results are achieved with Mixup.
We further show that gains hold for robustness to common input corruptions and perturbations at varying severities with a 0.7\% improvement on CIFAR-100-C, by combining StackMix with AugMix over AugMix.
On its own, improvements with StackMix hold across different number of labeled samples on CIFAR-100, maintaining approximately a 2\% gap in test accuracy --down to using only 5\% of the whole dataset-- and is effective in the semi-supervised setting with a 2\% improvement with the standard benchmark $\Pi$-model.
Finally, we perform an extensive ablation study to better understand the proposed algorithm.
\end{abstract}

\begin{figure}
  \centering
  \begin{subfigure}[b]{1\linewidth}
    \includegraphics[width=\linewidth]{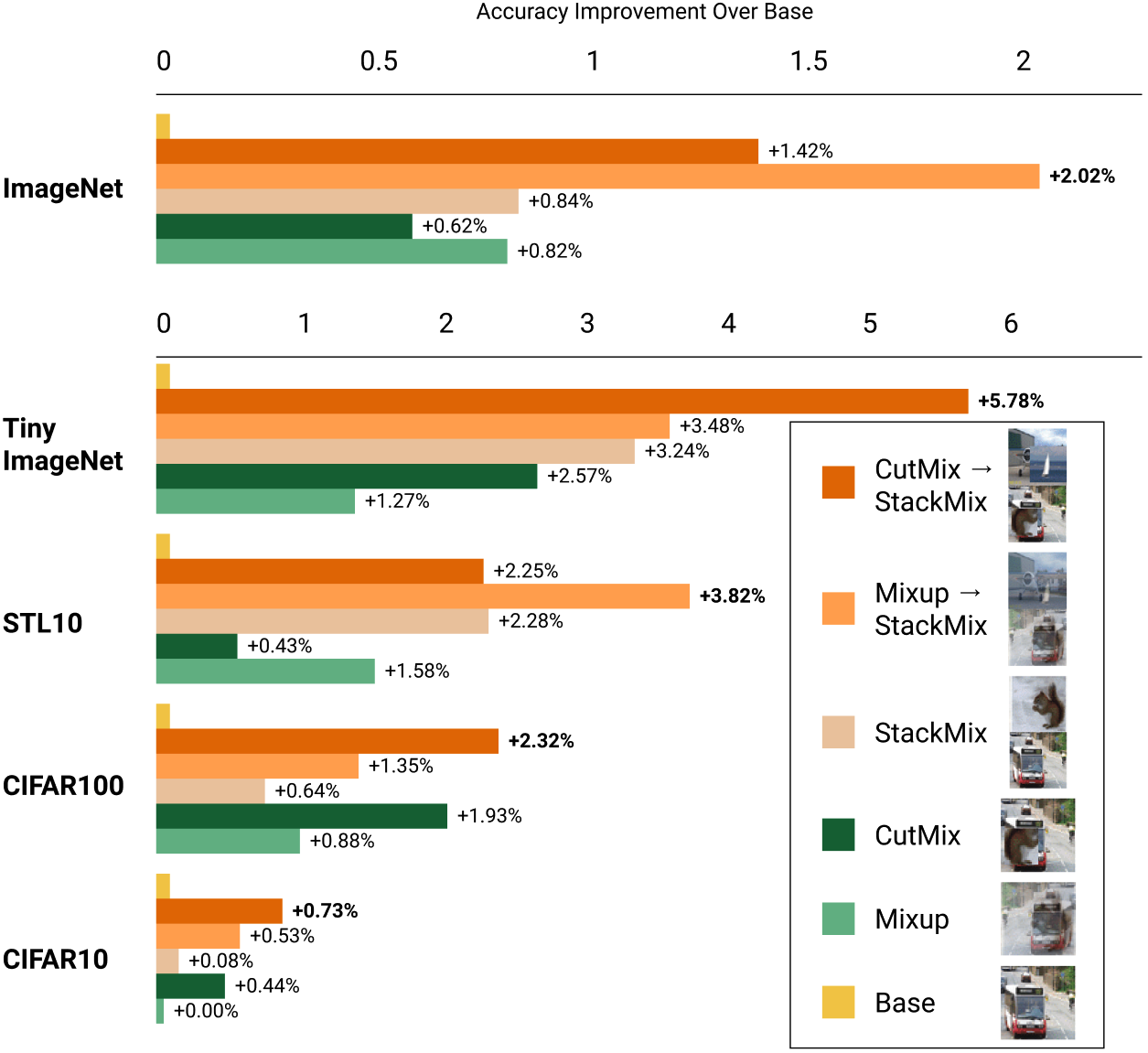}
    
  \end{subfigure} 
  \vspace{-0.8cm}
  \caption{\small Summary of results. Base is the typical data augmentation setting, with random crops, horizontal flips, and normalization. Accuracy improvement over base refers to the attained test set accuracy minus the base test set accuracy. A$\rightarrow$B refers to generating inputs with A and then feeding them as input to B, e.g. CutMix$\rightarrow$StackMix first generates inputs with CutMix and then feeds them as input to StackMix. StackMix variants perform by far the best, and exhibit complementary behavior with existing augmentation. MixUp$\rightarrow$CutMix and CutMix$\rightarrow$MixUp omitted due to worse performance than base (see Tables \ref{supervisedImageClassificationImageNet},\ref{supervisedImageClassification}). Exact results and settings given in Section \ref{resultsSection} and Tables \ref{supervisedImageClassificationImageNet},\ref{supervisedImageClassification}.
  } \label{fig:mainlineresults} \vspace{-0.7cm}
\end{figure}

\begin{figure*}
  \centering
  \begin{subfigure}[b]{1\linewidth}
    \includegraphics[width=\linewidth]{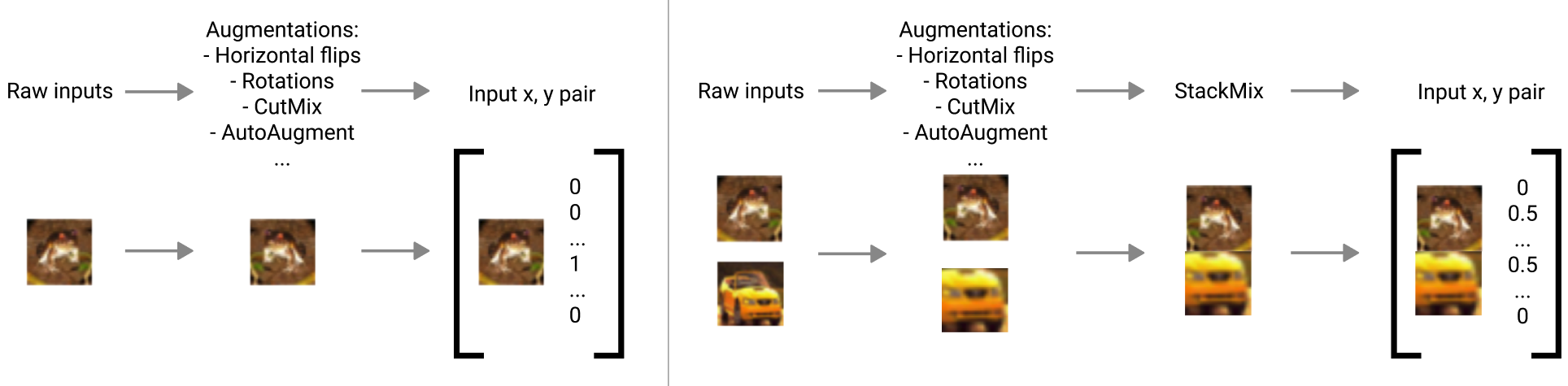}
    \label{fig:StackMix}
  \end{subfigure} 
  \vspace{-1cm}
  \caption{\small 
  The StackMix procedure. Left: The standard one-hot training. Right: StackMix with two images. Top: Abstract pipeline. Bottom: Concrete example. 
  }
  \label{fig:StackMix} \vspace{-0.4cm}
\end{figure*}

\section{Introduction}
In the last decade, numerous innovations in deep learning for computer vision have substantially improved results on many benchmark tasks \cite{krizhevsky2012imagenet, he2016deep, zagoruyko2016wide, huang2017densely}.
These innovations include architecture changes, training procedure improvements, data augmentation techniques, and regularization strategies, among many others. 
In particular, data augmentation techniques have consistently and predictably improved neural network performance, and remain crucial in training deep neural networks effectively. 

One such recent line of work revolves around the idea of finding effective augmentations through a search procedure \cite{cubuk2018autoaugment}. 
The resulting augmentations tend to outperform hand-designed algorithms \cite{cubuk2018autoaugment}, and have seen some adoption \cite{tan2020efficientnet, berthelot2019mixmatch}. 
There is work to reduce the cost of the search \cite{lim2019fast, ho2019population}.

A different line of work follows the idea of Mixup \cite{zhang2017mixup}, where inputs are generated from convex combinations of images and their labels. 
The resulting image can be understood as one image overlaid on another, with some opacity. 
Follow up works include methods such as Cutout \cite{devries2017improved} where parts of an image are removed, or CutMix \cite{yun2019cutmix} where parts of one image are removed and pasted onto another, with correspondingly weighted labels.
Other works further improve accuracy \cite{kim2020puzzle}, or robustness \cite{hendrycks2020augmix}. 
\emph{While highly effective individually, they generally cannot be combined with each other (See MixUp$\rightarrow$CutMix and CutMix$\rightarrow$MixUp in Tables \ref{supervisedImageClassificationImageNet},\ref{supervisedImageClassification}).} 
Furthermore, many cannot effectively combine more than 3 images (see Table \ref{tableAblation}), or they suffer from information loss due to inappropriate occlusion.


In this paper, we consider the supervised setting and introduce StackMix, a complementary Mix algorithm.
In StackMix, each input is presented as a concatenation of two images, and the label is the mean of the two one-hot labels.
We show StackMix works well with existing tuned hyperparameters, and has no change to existing losses or general network architecture, which allows for easy adoption and integration into modern deep learning pipelines. 
Most importantly, not only is StackMix an effective augmentation on its own, it can further boost the performance of existing data augmentation, including the Mix line of work.

Our contributions are as follows:\vspace{-0.2cm}
\begin{itemize}[leftmargin=*]
    \item We propose StackMix, a Mix data augmentation method that is complementary to existing augmentation.
    \vspace{-0.2cm}
    \item Compared to the vanilla case, StackMix improves the test performance on existing image classification tasks, including by 0.84\% on ImageNet with ResNet-50, 3.24\% on Tiny ImageNet with ResNet-56 \cite{he2016deep}, 1.30\% on CIFAR-100 with VGG-16 \cite{simonyan2014very} and 0.64\% with PreAct ResNet-18, 0.08\% on CIFAR-10 with SeResNet-18 and 0.14\% with ResNet-20, 2.28\% on STL-10 with Wide-ResNet 16-8 \cite{zagoruyko2016wide}. Finally, StackMix improves by 2.16\% on CIFAR-10, with all but 4000 labeled samples, when combined with the semi-supervised $\Pi$-model \cite{laine2017temporal}. \vspace{-0.2cm}
    \item We demonstrate that StackMix is complementary to existing data augmentation techniques, achieving over 0.8\% improvement on ImageNet, 3\% improvement on Tiny ImageNet, 2\% test error improvement on CIFAR-100, 0.5\% on CIFAR-10, and 1.5\% on STL-10, by combining StackMix with state-of-the-art data augmentation method CutMix \cite{yun2019cutmix}, as compared to CutMix alone. Similar gains are achieved with MixUp \cite{zhang2017mixup} and AutoAugment \cite{cubuk2018autoaugment}. In this way, many images are effectively combined. \vspace{-0.2cm}
    \item Improvements carry over to the robustness setting, with ~1\% test error improvement on CIFAR100-C \cite{hendrycks2019robustness} and 0.2\% on CIFAR10-C, by combining StackMix with state-of-the-art data augmentation method for robustness Augmix \cite{hendrycks2020augmix} over AugMix. \vspace{-0.2cm}
\end{itemize}


\section{The StackMix algorithm}
In StackMix, we alter the input to the network to be a concatenation of two images, and the output to be a two-hot vector of $\sfrac{1}{2}$ and $\sfrac{1}{2}$; see Figure \ref{fig:StackMix}. 
The choice of $\sfrac{1}{2}$ value is a result of using the Cross Entropy loss, and 1 and 1 can be explored for the Binary Cross Entropy loss. 
StackMix is tightly related to the line of ``Mix'' data augmentation work. StackMix has the following general advantages:\vspace{-0.2cm}
\begin{enumerate}[leftmargin=*,label=(\alph*)]
    \item StackMix is complementary to existing data augmentations including the ``Mix'' line of work (see Section \ref{complementary_to_aug}). This means StackMix can effectively mix more than two images, e.g. StackMix with $k=2$ and Mixup can effectively mix four images in total. This is in contrast to, for example, Mixup which does not benefit from mixing more than two images (see Page 3 of Mixup \cite{zhang2017mixup} or Table \ref{tableAblation}). In addition, the various ``Mix'' based methods cannot be effectively combined (See MixUp$\rightarrow$CutMix, CutMix$\rightarrow$MixUp in Tables \ref{supervisedImageClassificationImageNet},\ref{supervisedImageClassification}).  \vspace{-0.2cm}
    \item StackMix has no additional hyperparameters. \vspace{-0.2cm}
    \item Compared with methods which remove or replace parts of images, StackMix has no assumption that critical information is effectively captured in bounding boxes, which may not be the case for real-world datasets.  \vspace{-0.2cm}
\end{enumerate}

This construction can be directly plugged into any existing image classification training pipeline, with the only typical changes being the sizes of the first and last layers of the network. 
The change in parameters is generally insignificant (e.g., $<1\%$ for ResNet-20 on CIFAR10, or 0\% for PreAct ResNet-18 on CIFAR100, due to average pool; see Table \ref{tableModelParams} in Appendix; see Tables \ref{supervisedImageClassificationImageNet},\ref{supervisedImageClassification} for controls). 
To ensure fairness in comparisons, we tune hyperparameters in the original standard one-hot supervised setting --including epochs to ensure performance has saturated-- and we then apply the \textbf{exact same hyperparameters} to StackMix. 
We note that, for testing, we concatenate the same image twice, with the one-hot vector used as the ground truth label (see later section for discussion).

\subsection{Implementation and synergy with existing data augmentation}
In the traditional setting, a batch size of $k$ is defined by having $k$ inputs per batch, where each of the $k$ inputs is typically the result after data augmentation. 
For consistency with data augmentation techniques, which combine two or more images such as Mixup \cite{zhang2017mixup}, we define an input vector as a vector after the concatenation. 
In particular, and for simplicity of presentation, for each input, we assume we perform the following motions: \vspace{-0.2cm}

\begin{figure}[t]
\centering
\begin{minipage}{\linewidth}
    \vspace{-0.1cm}
    \begin{algorithm}[H]
        \caption{The StackMix algorithm. Produces one sample. For concatenating two images, we set $k=2$. To recover the standard one-hot supervised training, we set  $k=1$.}\label{alg:StackMix}
        \begin{algorithmic}
            \State \textbf{Inputs}: Samples $\left\{x_i, y_i\right\}_{i=0}^{k}$; $x_i$ are inputs and $y_i$ are one-hot labels; stochastic transformation $T$; number of images to concatenate $k$.  
            \State \quad 1. Compute $x_i = T(x_i)$.
            \State \quad 2. Concatenate as $x = \texttt{concat}\left(\{x_i\}_{i=0}^{k}\right)$ 
            \State \quad 3. Compute prediction $y = \tfrac{1}{k} \left(\sum_{i=0}^{k}{y_i}\right)$
            \State \Return $x, y$
        \end{algorithmic}
    \end{algorithm}
\end{minipage}
\end{figure}
\vspace{-0.1cm}
\begin{enumerate}[leftmargin=*,label=(\alph*)]
    \item Sample two images. \vspace{-0.2cm}
    \item Apply existing data augmentation to each image individually. \vspace{-0.2cm}
    \item Concatenate the two images as a single input vector. \vspace{-0.2cm}
    \item Rescale each label vector to $\sfrac{1}{2}$, and add them element-wise to produce the multi-hot label. \vspace{-0.2cm}
\end{enumerate}

This method can be easily extended to $k$-fold concatenation of images, where each label vector is rescaled to $\sfrac{1}{k}$, and then summed element-wise.
We explore $k>2$ in Section \ref{ablation}. 
For clarity, we present this procedure as well in Algorithm \ref{alg:StackMix}, where $k=1$ is the standard one-hot training procedure, and $k=2$ is the primary focus of this paper. In implementation, we sample two images with replacement and thus the output can be a one-hot vector, naturally with $\sfrac{1}{n}$ probability, where $n$ is the size of the dataset; although, we note that this choice has minimal impact on performance.

\section{Results} \label{resultsSection}
We provide experimental results for supervised image classification, test error robustness against image corruptions and perturbations
, semi-supervised learning, combining StackMix with existing data augmentation, an ablation study, and evaluation of test time augmentation. 
A summary of experimental settings are given in Table \ref{experimentSummary}, and comprehensively detailed in each section. 
\emph{We tuned the hyperparameters of the standard one-hot setting to achieve the performance of the original papers and of the most popular public implementations, reusing the most widely used codebases for consistency.}
We then used the \emph{exact same hyperparameters and pipeline} for StackMix for fairness.

\begin{table*}
\centering
\begin{small}
\begin{tabular}{lclclcl} 
\toprule
    Experiment short name & & Model & & Dataset & & Setting \\ 
    \cmidrule{1-1} \cmidrule{3-3} \cmidrule{5-5} \cmidrule{7-7}
    \texttt{RN50-IMAGENET} & & ResNet-50 & & ImageNet & & Supervised Learning \\
    \texttt{RN56-TINYIMAGENET} & & ResNet-56 & & Tiny ImageNet & & Supervised Learning \\
    \texttt{VGG16-CIFAR100} & & VGG-16 & & CIFAR100 & & Supervised Learning \\
    \texttt{PRN18-CIFAR100} & & PreActResNet-18 & & CIFAR100 & & Supervised Learning \\
    \texttt{SRN18-CIFAR10} & & SeResNet-18 & & CIFAR10 & & Supervised Learning \\
    \texttt{RN20-CIFAR10} & & ResNet-20 & & CIFAR10 & & Supervised Learning \\
    \texttt{WRN-STL10} & & Wide ResNet 16-8 & & STL10 & & Supervised Learning \\
    \texttt{WRN-CIFAR10-SSL} & & Wide ResNet 28-2 & & CIFAR10 & & Semi-Supervised Learning \\
    \texttt{WRN-CIFAR10/100-C} & & Wide ResNet 40-2 & & CIFAR10/100-C & & Robustness \\
    \texttt{RN20-CIFAR10-N} & & ResNet-20 & & CIFAR10 & & Ablation \\
    \texttt{VGG16-CIFAR100-N} & & VGG-16 & & CIFAR100 & & Ablation \\
    \texttt{PRN18-CIFAR100-INF} & & PreActResNet-18 & & CIFAR100 & & Test time inference augmentation \\
 \bottomrule
\end{tabular}\vspace{-0.1cm}
\caption{Summary of experimental settings. 
}\vspace{-0.1cm}
\label{experimentSummary}
\end{small}
\end{table*}

\subsection{Supervised Image Classification}
In this section, we explore improving the performance of well-known baselines in the supervised learning setting. 
We add StackMix to seven model-dataset pairs, and lastly observe the performance with and without StackMix across a varying number of supervised samples in the CIFAR100 setting. 
See Tables \ref{supervisedImageClassificationImageNet},\ref{supervisedImageClassification} for results.

\textbf{Controls.} Although StackMix generally introduces a small number of additional parameters (see Table \ref{tableModelParams} in Appendix), it is crucial to introduce controls to account for the difference, in addition to the increased training time. Therefore, we also present results with two controls. To account for the additional hyperparameters and computation, we introduce a control where the StackMix procedure concatenates the same image with itself during training, after being individually augmented with the stochastic image augmentation for fairness. To account for increased training time, we introduce a control with double the batch size and double the epochs, with re-tuned learning rate. This way we effectively control for both the model size and the total computation/number of images seen by the model during training. Results are presented in Tables \ref{supervisedImageClassificationImageNet},\ref{supervisedImageClassification} (See Base, 2x bs/epochs, StackMix(same)). 
It appears that neither control exhibits the same improvement as with StackMix. 
\textit{This is critical in our analysis of StackMix as this suggests the effect of StackMix is nontrivial and cannot be explained by computation or model size differences.}

\begin{figure*}
  \centering
    \begin{subfigure}[b]{0.30\linewidth}
    \includegraphics[width=\linewidth,trim={1.7cm 1.3cm 0 0},clip]{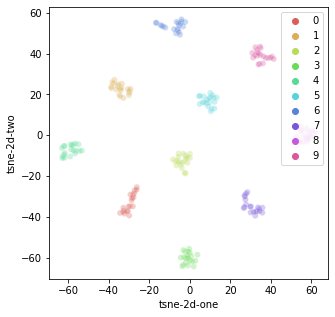}
    \label{fig:10classes}
  \end{subfigure}
    \begin{subfigure}[b]{0.30\linewidth}
    \includegraphics[width=\linewidth,trim={1.7cm 1.3cm 0 0},clip]{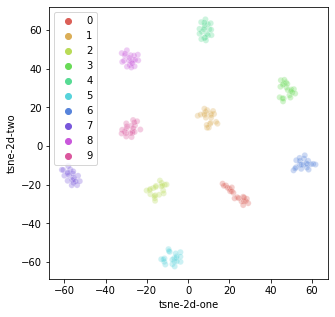}
    \label{fig:1class}
  \end{subfigure}
    \begin{subfigure}[b]{0.30\linewidth}
    \includegraphics[width=\linewidth,trim={1.7cm 1.3cm 0 0},clip]{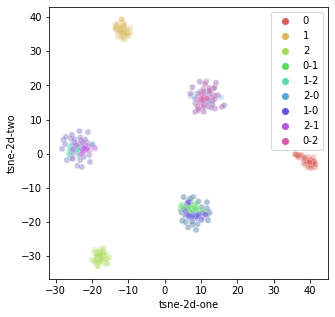}
    \label{fig:3classes}
  \end{subfigure}
  \vspace{-0.3cm}
  \caption{\small Left: Randomly selected 10 classes where each image is concatenated with itself. Middle: Randomly selected 10 classes where each image is concatenated with an image from class 0. Right: Randomly selected 3 classes where each image is concatenated with itself, concatenated as the top image with an image from another class, and concatenated as the bottom image with an image from another class. Singular number denotes self-concatenation. ``a-b'' denotes image from class a concatenated as the top image with an image from class b. Data used is training data and test data plots are similar with more noise. 
  }
  \label{fig:plotTSNE} \vspace{-0.3cm}
\end{figure*}

\textbf{Examining learned embeddings.} We check the learned embeddings for randomly drawn samples from CIFAR100 with t-SNE \cite{tsne}, given in Figure \ref{fig:plotTSNE}, as sanity check. The images are processed in the inference setting, where they are concatenated with themselves. Clusters form as expected (Figure \ref{fig:plotTSNE} Left). This is highly encouraging despite the network mostly seeing the concatenation of images from different classes. We also observe that by fixing one image of each concatenation to be a certain class and varying the class of the other image, a similarly separated distribution forms (Figure \ref{fig:plotTSNE} Middle). This further supports the idea that the network has learned to differentiate between the two presented images. Finally, we find that concatenating images from two different classes is semantically separated from concatenating either image with itself (Figure \ref{fig:plotTSNE} Right), and that as a sanity check the embeddings are generally not sensitive to which image is placed on top. In sum, while the network sees the same image stacked during testing and largely sees different images stacked during training, it appears to learn reasonable embeddings.

\begin{table}
\centering
\begin{small}
\begin{tabular}{c|c}
\toprule
    Method & Test error \\ \midrule
    Base & 24.10 \\
    2x bs/epochs & 24.49 \\
    StackMix(same) & 23.28 \\
    MixUp & 23.28 \\ 
    CutMix & 23.48 \\
    StackMix & 23.26 \\
    MixUp$\rightarrow$CutMix & 35.70 \\
    CutMix$\rightarrow$MixUp & 33.99 \\
    MixUp$\rightarrow$StackMix & \textbf{22.08} \\
    CutMix$\rightarrow$StackMix & 22.68 \\
 \bottomrule
\end{tabular}\vspace{-0.1cm}
\caption{Generalization error of experiments with and without StackMix in the \texttt{RN50-ImageNet} setting. 2x bs/epochs refers to doubling the batch size and epochs of Base. StackMix(same) is another control which refers to stacking the same image as input. A$\rightarrow$B refers to generating inputs with A and then feeding them as input to B, e.g. CutMix$\rightarrow$MixUp first generates inputs with CutMix and then feeds them as input to MixUp.
}\vspace{-0.1cm}
\label{supervisedImageClassificationImageNet}
\end{small}
\end{table}

\begingroup
\setlength{\tabcolsep}{2pt} 
\begin{table*}
\centering
\begin{scriptsize}
\begin{tabular}{ccccccccccc}
\toprule
    Experiment & Base & 2x bs/epochs & StackMix(same) & MixUp & CutMix & StackMix & MixUp$\rightarrow$CutMix & CutMix$\rightarrow$MixUp & MixUp$\rightarrow$StackMix & CutMix$\rightarrow$StackMix \\ 
    \midrule
    \texttt{RN56-TINYIMAGENET} & 42.03 & 42.00 & 42.11  & 40.76 & 39.46 & 38.79 & 43.80 & 43.24 &  38.55 & \textbf{36.25} \\
    \texttt{VGG16-CIFAR100} & 27.80 & 28.63 & 27.69  & 27.35 & 27.20 &  26.50 & 34.44 & 35.66 &  25.69 & \textbf{25.49} \\
    \texttt{PRN18-CIFAR100} & 25.93 & 25.41 & 25.63  & 25.05 & 24.00 & 25.29 & 31.36 & 30.20 &  24.58 & \textbf{23.61} \\
    \texttt{RN20-CIFAR10} & 7.65 & 7.55 & 7.73  & 6.51 & 6.74 & 7.51 & 6.93 & 7.05 &  6.40 & \textbf{6.27} \\
    \texttt{SRN18-CIFAR10} & 5.03 & 5.05 & 5.21  & 5.34 & 4.59 & 4.95 & 6.64 & 6.59 &  4.50 & \textbf{4.30} \\
    \texttt{WRN-STL10} & 17.26 & 15.83 & 18.92  & 15.68 & 16.83 & 14.98 & 24.02 & 23.68 &   \textbf{13.44} & 15.01 \\
 \bottomrule
\end{tabular}\vspace{-0.1cm}
\caption{Generalization error of experiments with and without StackMix in the supervised setting. 2x bs/epochs refers to doubling the batch size and epochs of Base. StackMix(same) is another control which refers to stacking the same image as input. A$\rightarrow$B refers to generating inputs with A and then feeding them as input to B, e.g. CutMix$\rightarrow$MixUp first generates inputs with CutMix and then feeds them as input to MixUp.
}\vspace{-0.1cm}
\label{supervisedImageClassification}
\end{scriptsize}
\end{table*}
\endgroup

\begin{table*}
    \centering
    \small
    \begin{tabular}{ccccccc}
    \toprule
        samples\% & 100 & 50 & 30 & 20 & 10 & 5  \\\midrule
        Base & 27.80\% $\pm$.10 & 34.88\% $\pm$.20 & 42.52\% $\pm$.34 & 50.41\% $\pm$.38 & 71.91\% $\pm$.57 & 86.03\% $\pm$.12 \\
        StackMix & \textbf{26.50\% $\pm$.11} & \textbf{33.61\%} $\pm$.21 & 
        \textbf{40.40\%} $\pm$.34 & \textbf{48.19\%} $\pm$.52 & \textbf{68.71\%} $\pm$.87 & \textbf{85.61\%} $\pm$.40 \\ \bottomrule
    \end{tabular} \vspace{-0.1cm}
    \caption{Generalization error (\%) for VGG16-CIFAR100 with varying number of proportional samples in each class.}
    \label{vgg16Cifar100VariousEpochsTable}\vspace{-0.1cm}
\end{table*}

\textbf{\texttt{ImageNet}.} ImageNet \cite{russakovsky2015imagenet} is a dataset with 1.3 million training images and 50,000 validation images of varying dimensions and 1,000 classes.
ResNet-50 \cite{he2016deep} is a deep ResNet architecture with 50 layers. We use the official PyTorch implementation and train the network for the default 90 epochs, which roughly follows popular works \cite{he2016deep, huang2017densely, han2017deep, simonyan2014very, zhang2017mixup}. There are some works which train the network for 3-4x the number of epochs, e.g. CutMix \cite{yun2019cutmix}, but this can be computationally demanding. We use the standard random crops of size $224\times224$, horizontal flips, and normalization. In inference, we use a $224\times224$ center crop, following standard. The network is trained with momentum SGD ($\eta = 0.1$, $\beta=0.9$), with a 30-60 decay schedule by factor of 0.1 using a batch size of 256. We set $\alpha=1$ for MixUp and CutMix. 
StackMix variants perform by far the best, with best StackMix variant improving $2.02\%$ over base. Adding StackMix to MixUp improves $1.20\%$ over MixUp, and adding StackMix to CutMix improves $0.80\%$ over CutMix (see Table \ref{supervisedImageClassificationImageNet} and Figure \ref{fig:mainlineresults}).

\begin{figure}
  \centering
    \begin{subfigure}[b]{0.49\linewidth}
    \includegraphics[width=\linewidth]{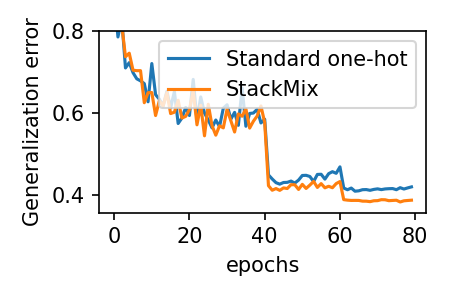}
    \label{fig:rn56tinyimagenet}
  \end{subfigure}
  \begin{subfigure}[b]{0.49\linewidth}
    \includegraphics[width=\linewidth]{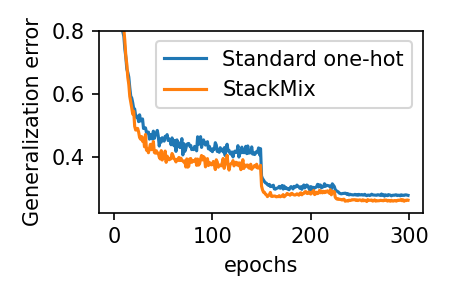}
    \label{fig:vgg16cifar100}
  \end{subfigure} 
  \vspace{-0.7cm}
  \caption{\small Generalization error for supervised learning. Left: \texttt{RN56-TINYIMAGENET}. Right: \texttt{VGG16-CIFAR100}. 
  }
  \label{fig:plots1}\vspace{-0.5cm} 
\end{figure}

\textbf{\texttt{Tiny ImageNet}.} Tiny ImageNet has 110,000 images of size $64\times64\times3$ and 200 classes. 
The test/train split is 100,000/10,000. With ResNet-56 \cite{he2016deep}, we trained the model for 80 epochs with momentum SGD ($\eta = 0.1$, $\beta=0.9$), Cross Entropy loss, decaying by a factor of 0.1 at 40 and 60 epochs, using a batch size of 64. 
We applied the standard image augmentation \cite{he2016deep} of horizontal flips, normalization and random crops. 
By adding StackMix to the vanilla case, the absolute generalization error was reduced by 3.24\%, from 42.03\% to 38.79\%. 
By observing Figure \ref{fig:plots1} (Left), we see that while the two methods are initially comparable, adding StackMix reduces the error in the later stages of training. 
The plateau of the StackMix curve suggests resistance to overfitting. Furthermore, by adding StackMix to MixUp, test error is decreased by 2\%, and by adding StackMix to CutMix, test error is decreased by 3\% (see Table \ref{supervisedImageClassification} and Figure \ref{fig:mainlineresults}).

\textbf{\texttt{CIFAR100}.} CIFAR100 has 100 classes and 500 samples per class in the training set, and 100 samples per class in the test set. 
Images are of size $32\times32\times3$. We trained two models, VGG16 and PreActResNet-18 (PRN18). VGG-16 was trained for 300 epochs following standard procedure as in \texttt{Tiny ImageNet}. PRN18 was trained similarly, except for 200 epochs and a learning rate decay schedule by a factor 0.2 at 60, 120, and 180 epochs.  A roughly 1\% test error improvement is observed for both cases for StackMix compared to the controls. Relative to MixUp and CutMix, a significant decrease of 1-2\% is observed by adding StackMix (see Table \ref{supervisedImageClassification} and Figure \ref{fig:mainlineresults}). Contrary to \texttt{Tiny ImageNet}, we observe in Figure \ref{fig:plots1} (Right) that StackMix already improves in the early stages of training with VGG16.  It is typical in neural network training to see the gap closed in the first learning rate decay when there exists a gap early on in training, but here StackMix maintains an improvement.

\textbf{\texttt{CIFAR10}.} CIFAR10 is the 10 class version of CIFAR100. We trained two models, ResNet20 (RN20) and SeResNet-18 (SRN18). This is a particularly challenging task to improve upon due to the model architecture where doubling the number the parameters and increasing the depth results in only minor gains in performance \cite{he2016deep}. Both networks are trained similarly as previously, except with a 30-60-90 learning rate decay schedule for SRN18. In both cases there are small improvements over the controls, and adding StackMix to existing augmentation further improves results.
We emphasize that ResNet20 is not ResNet18, which is a different network architecture with significantly more parameters. Most popular implementations of ResNet20 fall in the 8-8.5\% test error range on CIFAR10. SeResNet18 is a variation of ResNet18.

\begingroup
\setlength{\tabcolsep}{3pt} 
\begin{table*}
\centering
\begin{scriptsize}
\begin{tabular}{c|c|ccc|cccc|cccc|cccc|c}
\toprule
    & & \multicolumn{3}{c}{Noise} & \multicolumn{4}{c}{Blur} & \multicolumn{4}{c}{Weather} & \multicolumn{4}{c}{Digital} & \\ \midrule
    Setting & Clean & Gauss. & Shot & Impulse & Defocus & Glass & Motion & Zoom & Snow & Frost & Fog & Bright & Contrast & Elastic & Pixel & JPEG & \textbf{mCE} \\ 
    \midrule
    \texttt{Standard} & 25.18 & 83.1 & 75.2 & 75.8 & 43.2 & 78.4 & 49.6 & 50.4 & 48.3 & 53.5 & 39.6 & 30.0 & 48.1 & 44.1 & 54.4 & 55.0 & 55.24 \\
    \texttt{StackMix} & 24.80 & 81.4 & 73.8 & 74.0 & 44.0 & 77.5 & 49.8 & 51.3 & 46.7 & 52.8 & 37.6 & 29.6 & 47.7 & 44.1 & 55.7 & 55.7 & 54.78 \\
    \texttt{AugMix} & 23.37 & \textbf{54.5} & \textbf{46.7} & 36.7 & 25.6 & \textbf{53.1} & 29.6 & 28.2 & 34.7 & 36.0 & 33.0 & 25.6 & 31.3 & 32.4 & \textbf{36.5} & \textbf{38.3} & 36.14 \\
    \texttt{AugMix$\rightarrow$StackMix} & \textbf{21.81} & 55.7 & 47.3 & \textbf{33.0} & \textbf{24.0} & 54.9 & \textbf{27.8} & \textbf{27.2} & \textbf{32.8} & \textbf{34.9} & \textbf{29.9} & \textbf{23.9} & \textbf{30.9} & \textbf{31.3} & 39.3 & 39.0 & \textbf{35.46} \\

 \bottomrule
\end{tabular}\vspace{-0.1cm}
\caption{Generalization error of experiments with and without StackMix in the \texttt{WRN-CIFAR100-C} setting. }\vspace{-0.1cm}
\label{robustnessMCECIFAR100}
\end{scriptsize}
\end{table*}
\endgroup

\begingroup
\setlength{\tabcolsep}{3pt} 
\begin{table*}
\centering
\begin{scriptsize}
\begin{tabular}{c|c|ccc|cccc|cccc|cccc|c}
\toprule
    & & \multicolumn{3}{c}{Noise} & \multicolumn{4}{c}{Blur} & \multicolumn{4}{c}{Weather} & \multicolumn{4}{c}{Digital} & \\ \midrule
    Setting & Clean & Gauss. & Shot & Impulse & Defocus & Glass & Motion & Zoom & Snow & Frost & Fog & Bright & Contrast & Elastic & Pixel & JPEG & \textbf{mCE} \\ 
    \midrule
    \texttt{Standard} & 5.49 & 51.1 & 39.1 & 43.1 & 19.1 & 50.5 & 24.2 & 25.2 & 19.1 & 23.2 & 11.9 & 7.1 & 21.8 & 16.7 & 30.0 & 22.6 & 26.98 \\
    \texttt{StackMix} & 5.30 & 53.0 & 42.5 & 43.1 & 18.2 & 49.4 & 23.6 & 23.8 & 18.5 & 21.7 & 11.7 & 6.8 & 22.3 & 17.5 & 29.7 & 23.0 & 26.99 \\
    \texttt{AugMix} & 4.91 & \textbf{22.2} & \textbf{16.3} & 13.0 & 5.8 & \textbf{21.0} & 7.9 & 7.2 & 10.7 & 10.4 & 8.2 & 5.6 & 7.8 & 10.1 & 16.3 & 12.6 & 11.67 \\
    \texttt{AugMix$\rightarrow$StackMix} & \textbf{4.37} & 24.2 & 16.8 & \textbf{10.7} & \textbf{5.3} & 22.4 & \textbf{7.5} & \textbf{6.7} & \textbf{9.9} & \textbf{10.0} & \textbf{7.5} & \textbf{5.0} & \textbf{7.3} & \textbf{9.6} & \textbf{16.2} & \textbf{12.5} & \textbf{11.44} \\

 \bottomrule
\end{tabular}\vspace{-0.1cm}
\caption{Generalization error of experiments with and without StackMix in the \texttt{WRN-CIFAR10-C} setting.}\vspace{-0.1cm}
\label{robustnessMCECIFAR10}
\end{scriptsize}
\end{table*}
\endgroup
\begin{table}
\centering
\begin{small}
\begin{tabular}{lclclcl} 
\toprule
    Experiment & & Base & & StackMix\\ 
    \cmidrule{1-1} \cmidrule{3-3} \cmidrule{5-5} 
    \texttt{WRN-CIFAR10-SSL} & & 17.31\% & & \textbf{15.15}\%\\ \bottomrule
\end{tabular}\vspace{-0.1cm}
\caption{Generalization error of $\Pi$-model on the standard benchmark of CIFAR10, with all but 4,000 labels removed.}\vspace{-0.3cm}
\label{semisupervisedImageClassification}
\end{small}
\end{table}

\textbf{\texttt{STL10}.} 
STL-10 comprises 1300 images of size $96\times96\times 3$ with a 500/800 train/test split and 10 classes. 
This is challenging dataset due to the number of training samples and size of images. 
We use Wide ResNet 16-8 \cite{zagoruyko2016wide}, a 16 layer deep ResNet architecture with 8 times the width. 
We trained the WRN model for 100 epochs following standard settings as before. 
Similar results are observed in Table \ref{supervisedImageClassification}, except CutMix does not perform as well, leading to MixUp$\rightarrow$StackMix attaining the best performance. 

\textbf{Understanding performance with varying labeled samples.} 
StackMix performs well in the above supervised settings, and we further explore performance in the low sample regime. 
In particular, we select the \texttt{VGG16-CIFAR100} setting, and decrease the number of samples in each class proportionally. 
We use the exact same training setup as in the full \texttt{VGG16-CIFAR100} case, and tabulate results in Table \ref{vgg16Cifar100VariousEpochsTable}. 
We perform 3 runs since low-sample settings produce higher variance results. 
The improvement for the full dataset setting hold with lower samples at roughly 2\% generalization error.

\begin{table*}
\centering
\begin{small}
\begin{tabular}{ccccc} 
\toprule
    Experiment & Base & AA & StackMix & AA$\rightarrow$StackMix  \\ \midrule
    \texttt{PRN18-CIFAR100-AA} & 25.93\% & 23.87\% & 25.29\% &  \textbf{21.51\%} \\ \bottomrule
    
\end{tabular}\vspace{-0.1cm}
\caption{Generalization error (\%) of \texttt{PRN18-CIFAR100} with AutoAugment. It is no surprise that StackMix is complementary to AutoAugment, and we simply present one experiment here to confirm. }\vspace{-0.1cm}
\label{aaImageClassification}
\end{small}
\end{table*}
\begin{table*}
    \centering
    \small
    \begin{tabular}{ccccccccccccc} 
    \toprule
        & & & & \multicolumn{9}{c}{$k$}\\ \midrule
        Dataset-Model & & Augmentation & & 1 (base) & & 2 (same image) & & 2 & & 3 & & 5  \\ 
        \cmidrule{1-1} \cmidrule{3-3} \cmidrule{5-5} \cmidrule{7-7} \cmidrule{9-9} \cmidrule{11-11} \cmidrule{13-13}
        \multirow{5}{*}{\texttt{VGG16-CIFAR100}} & & Standard & & 27.80\% & & 27.69\% & & \textbf{26.50\%} & & 27.35\% & & 29.35\% \\
        && MixUp & & - & & - & & \textbf{27.35}\% & & 49.04\% & & 63.52\%\\
        && CutMix & & - & & - & & \textbf{27.20}\% & & 36.52\% & & 51.25\% \\
         && MixUp$\rightarrow$StackMix & & - & & - & & \textbf{25.69}\% & & 26.05\%  & & 27.41\%\\
         && CutMix$\rightarrow$StackMix & & - & & - & & \textbf{25.49}\% & & 26.95\%  & & 27.91\%\\
         \midrule
        \multirow{5}{*}{\texttt{RN-CIFAR10}} && StackMix & & 7.65\% & & 7.73\% & &  \textbf{7.51\%} & & 8.13\% & & 7.89\% \\
        && MixUp & & - & & - & & \textbf{6.51}\% & & 8.93\% & & 13.86\% \\
        && CutMix & & - & & - & & \textbf{6.74}\% & & 7.63\% & & 9.91\% \\
        && MixUp$\rightarrow$StackMix & & - & & - & & 6.40\% & & 6.30\% & & \textbf{6.05\%}\\
        && CutMix$\rightarrow$StackMix & & - & & - & & \textbf{6.27\%} & & 6.65\% & & 6.81\% \\
        \bottomrule
    \end{tabular} \vspace{-0.1cm}
    \caption{Generalization error for \texttt{VGG16-CIFAR100} and \texttt{RN20-CIFAR10} with varying number of images concatenated.}
    \label{tableAblation}\vspace{-0.4cm}
\end{table*}

\subsection{Robustness}
\textbf{\texttt{CIFAR10/100-C}.}  We investigate the impact of StackMix on robustness. 
In particular, we select a corrupted dataset as test set and reevaluate models trained with and without StackMix on the uncorrupted training set, following standardized procedure \cite{hendrycks2019robustness}. AugMix \cite{hendrycks2020augmix} follows the Mix line of work with significant increases in robustness, and therefore we investigate the state-of-the-art method here. Results with WRN-40-2 on CIFAR100-C and CIFAR10-C are shown in Tables \ref{robustnessMCECIFAR100} and \ref{robustnessMCECIFAR10},  respectively.

In the clean case, StackMix improves over the vanilla case, and adding StackMix to AugMix significantly decreases test error, including 2\% on CIFAR100. StackMix does not appear to provide any additional robustness improvements beyond improvements carried over from the clean case. However, this should not be taken for granted as some methods which increase robustness can lower clean test error and vice versa \cite{raghunathan2019adversarial}.
AugMix$\rightarrow$StackMix outperforms AugMix in both clean and corrupted cases on average, and outperforms AugMix in 11/15 categories in CIFAR10-C and 10/15 categories in CIFAR100-C.

\subsection{Semi-supervised Learning}
Thus far, StackMix is applied under the Cross Entropy loss. 
While varying the number of samples is helpful in understanding the impact of StackMix under different settings, we explore if StackMix can be directly applied to improve Semi-Supervised Learning (SSL) \cite{chapelle2006semi}, where the network processes both labeled and unlabeled samples. 
We select a popular and practical subset of SSL, which involves adding a loss function for consistency regularization \cite{tarvainen2017mean, berthelot2019mixmatch, laine2017temporal, chen2020negative}. Consistency regularization is similar to contrastive learning in that it tries to minimize the difference in output between similar samples. 
In particular, we select the classic and standard benchmark of the $\Pi$-model \cite{laine2017temporal}. 

The $\Pi$ model adds a loss function for the unlabeled samples of following form:\vspace{-0.2cm}
\begin{align}
d(f_{\theta}(x),f_{\theta}(\hat{x})), \nonumber 
\end{align}
where $d$ is typically the Mean Square Error, $f_{\theta}$ is the output of the neural network, and $\hat{x}$ is a stochastic perturbation of $x$. 
Minimizing this loss enforces similar output distributions of an image and its perturbation. 

A coefficient is then applied to the SSL loss as a weight with respect to the Cross Entropy loss. 
The unlabeled samples are evaluated with the SSL loss, while the labeled samples are evaluated with Cross Entropy.

\textbf{\texttt{CIFAR10}.} 
We follow the standard setup in \cite{oliver2018realistic} for the CIFAR10 dataset, where 4000 labeled samples are selected, and remaining samples are unlabeled. 
We use a WRN 28-2 architecture \cite{zagoruyko2016wide}, training for 200,000 iterations with a batch size of 200, of which 100 are labeled and 100 are unlabeled. 
The Adam optimizer is used ($\eta=3\cdot 10^{-4}, \beta_1=0.9, \beta_2=0.999$), decaying learning rate schedule by a factor of 0.2 at 130,000 iterations. 
Horizontal flips, random crops, and gaussian noise are used as data augmentation. 
A coefficient of 20 is used for the SSL loss. 
By adding StackMix, we reduce the test error by 2.16\% (see Table \ref{semisupervisedImageClassification}).

\subsection{StackMix is complementary to existing data augmentation} \label{complementary_to_aug}
Data augmentation is critical in training neural network models. 
Recently, stronger forms of data augmentation \cite{zhang2017mixup, devries2017improved, yun2019cutmix, cubuk2018autoaugment} have substantially improved results. 
Results in the previous section strongly suggest that StackMix is complementary to Mix methods, with improved training by combining StackMix with MixUp \cite{zhang2017mixup}, CutMix \cite{yun2019cutmix} and AugMix \cite{hendrycks2020augmix}. This differs from some existing work, for example MixUp and CutMix cannot be effectively combined (see Tables \ref{supervisedImageClassificationImageNet},\ref{supervisedImageClassification}). We further support the complementary nature of StackMix by combining with AutoAugment \cite{cubuk2018autoaugment}.


\textbf{\texttt{CIFAR100}.} We follow experimental settings in \texttt{PRN18-CIFAR100}. We use existing AutoAugment policies for the CIFAR datasets, and following \cite{cubuk2018autoaugment} for CIFAR, we apply AutoAugment after other augmentations, and before normalization and StackMix. AutoAugment improves 2\% over standard augmentation, and adding StackMix improves by another 2\% (see Table \ref{aaImageClassification}); again, suggesting a complementary behavior and easy incorporation into existing pipelines. We want to emphasize the result in this section. \textit{A 2\% gain by combining StackMix with AutoAugment over either baseline on the CIFAR100 dataset is comparable to a significant increase in model size and depth; on CIFAR100, typically moving from a ResNet18 model to ResNet101 and beyond on yields a roughly 2\% improvement in most implementations. }

\begin{figure*}
  \centering
    \begin{subfigure}[b]{0.45\linewidth}
    \includegraphics[width=\linewidth]{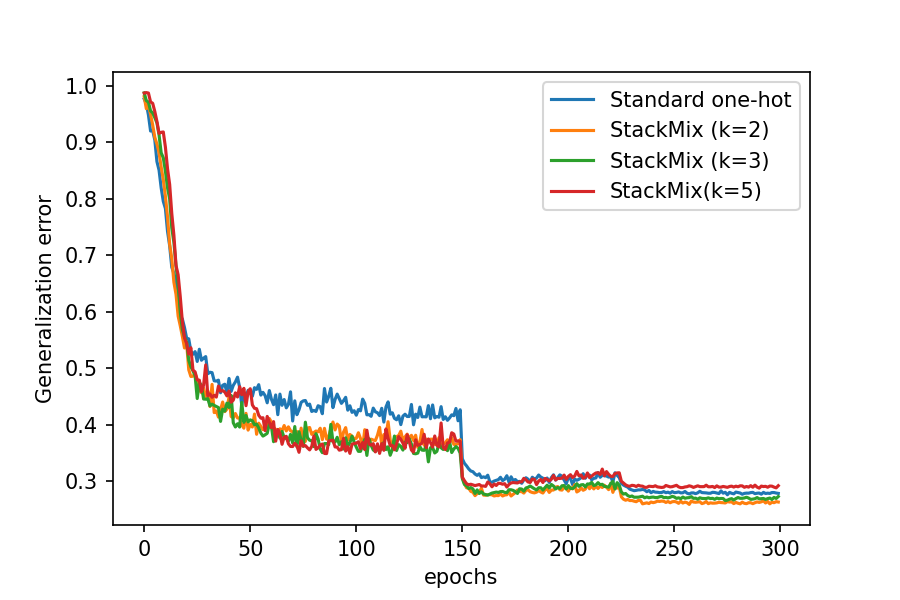}
    \label{fig:ablationvgg16}
  \end{subfigure}
    \begin{subfigure}[b]{0.45\linewidth}
    \includegraphics[width=\linewidth]{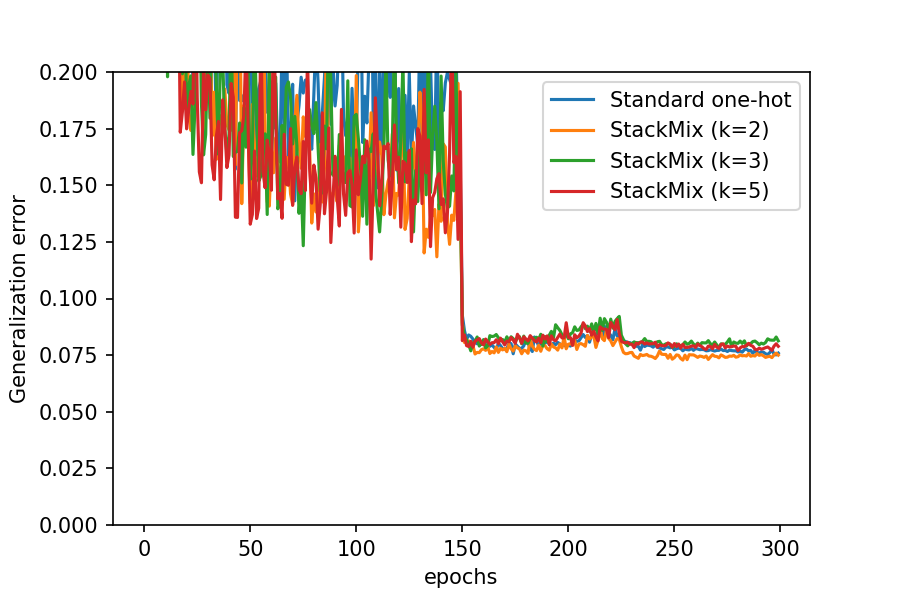}
    \label{fig:ablationrn20}
  \end{subfigure}

  \vspace{-0.7cm}
  \caption{\small Left: Generalization error for \texttt{VGG16-CIFAR100}. Right: Generalization error for \texttt{RN20-CIFAR10}. Varying number of images concatenated. 
  }
  \label{fig:plotAblation} \vspace{-0.3cm}
\end{figure*}
\begin{table*}
\centering
\begin{small}
\begin{tabular}{ccccc} 
\toprule
    Experiment & Base & Base + flips & StackMix & StackMix + flips  \\ \midrule
    \texttt{PRN18-CIFAR100-INF} & 25.93\% & 25.36\% & 25.29\% & 24.79\% \\ \bottomrule
    
\end{tabular}\vspace{-0.1cm}
\caption{Generalization error (\%) of \texttt{PRN18-CIFAR100} with test time augmentation. Base + flips is the mean output of an image and its flipped version. StackMix + flips is the output of the concatenation of an image with itself flipped.}\vspace{-0.4cm}
\label{tableInference}
\end{small}
\end{table*}
\vspace{-0.1cm}
\subsection{Ablation study} \label{ablation}
In this paper, we studied StackMix in the setting of two images stacked ($k=2$ in Algorithm \ref{alg:StackMix}). 
We now perform an ablation study to determine how far this framework can be pushed. 
We increase the value of $k$, and observe the test error in the setting of \texttt{VGG16-CIFAR100} and \texttt{RN20-CIFAR100}. 
We fix the hyperparameters as used previously, with results in Table \ref{tableAblation} and Figure \ref{fig:plotAblation}. For MixUp and CutMix, $k$ represents the number of images combined. For MixUp/CutMix$\rightarrow$StackMix, $k$ represents the number of images stacked, after they have been pairwise augmented with MixUp/CutMix (e.g. $k=3$ would be $2\times3=6$ total). We reduce the box size of CutMix to allow for higher $k$.

In almost all cases, the error deteriorates immediately after $k=2$, and further increasing $k$ typically increases the error further, clearer in the case of \texttt{VGG16-CIFAR100}. Performance deterioration is significantly more severe for MixUp and CutMix, whereas the StackMix variants suffer only slightly. This is likely due to loss of semantic information with inputs looking similar for MixUp, and a failure to capture enough critical information for CutMix. For example, on CIFAR100 MixUp and CutMix almost double in error from $k=2$ to $5$, tenfold the error rate increase of the StackMix variants.
We can see in Figure \ref{fig:plotAblation} that the choice of $k$ has limited impact in the early stages of training, but affects the final test error, where performance begins to deteriorate after the first learning rate decay. 

Furthermore, we highlight results on the concatenation of the same image (also in Tables \ref{supervisedImageClassificationImageNet},\ref{supervisedImageClassification}). 
First, this results in a sanity check that the StackMix construction on the same image is identical (with respect to performance) to the one-hot vector classification constructions.
Second, worse performance in StackMix when the same image is concatenated twice indicates that the network learns less, as compared to the concatenation of two images: this further strengthens the effect that StackMix brings during training.

\subsection{Inference speed and augmentations}
One drawback of StackMix compared with the standard one-hot is slower inference speed due to the larger input size. Therefore, we design an experiment where the standard one-hot case is given two forward passes for inference at test time. Concretely, we take the top-1 of the mean output of an image and its flipped counterpart. For StackMix in this paper, we concatenated the same image with itself without any further augmentation. However, we observe that the benefits of test-time augmentation for the standard case carry over to StackMix naturally without additional computation, where an image can be concatenated with a flipped version of itself. See Table \ref{tableInference} for results. The improvements with respect to each vanilla case are similar, where the standard case gains 0.57\% and StackMix gains 0.50\%.

\vspace{-0.1cm}
\section{Related Work}
In supervised learning, several ideas have been recently introduced that significantly boosts the performance in supervised learning.
These techniques can be added to the label, such as label smoothing \cite{sukhbaatar2014training}, or directly to the data, using data augmentation \cite{zhang2017colorful, cubuk2018autoaugment, devries2017improved, yun2019cutmix}, or both \cite{zhang2017mixup}. 

Horizontal image flips and cropping have been well-established as effective data augmentation techniques \cite{krizhevsky2012imagenet, he2016deep}. In one line of work, the choice of single-image augmentations was discovered through a search procedure \cite{cubuk2018autoaugment}. The cost of the method was reduced in further work \cite{ho2019population, lim2019fast}.

StackMix is tightly related to the ``Mix'' line of work \cite{zhang2017mixup, yun2019cutmix, devries2017improved, hendrycks2020augmix, kim2020puzzle, guo2018mixup}, where pairs of input images and their labels are combined. Mixup \cite{zhang2017mixup} takes convex combinations of inputs and their labels, justified under Occam's Razor. This idea has been extended to the feature space \cite{verma2019manifold}. Other work, such as Cutout, removes parts of images \cite{devries2017improved,zhong2017random}. Further extensions, such as CutMix,  removes and pastes parts of images with weighted labels \cite{yun2019cutmix, Takahashi_2020}. PuzzleMix \cite{kim2020puzzle} improves the salient information in Mix images, while AugMix \cite{hendrycks2020augmix} is another extension with improved augmentations for robustness. 

There are two related frameworks that output multiple labels from a single image, namely ensembles \cite{ensemble} and multiple choice learning \cite{guzman2012multiple}.
Both ensembles and multiple choice learning aim to output multiple labels from the same input; ensembles utilize multiple models to obtain multiple predictions from the same input, while multiple choice learning predicts multiple labels from the same model.
Recent literature in ensemble learning have explored improving an ensemble of neural networks \cite{hansen1990neural} with random initialization \cite{lakshminarayanan2017simple}, attention \cite{abe}, information theoretic objectives \cite{dibs}, among others.
StackMix is strictly different from both ensembles and multiple choice learning as our aim is to predict multiple outputs from multiple inputs.

\vspace{-0.1cm}
\section{Conclusion}\vspace{-0.1cm}
We introduce StackMix, a complementary Mix algorithm. 
StackMix can directly be plugged into existing pipelines with minimal changes: no change in loss, hyperparameters, or general network architecture. 
StackMix improves performance in standard benchmarks including ImageNet, Tiny ImageNet, CIFAR-10, CIFAR-100, and STL-10. 
Furthermore, StackMix improves robustness, and semi-supervised learning. StackMix is complementary to and boosts the performance of existing augmentation, including MixUp, CutMix, AugMix, and AutoAugment. 

{\small
\bibliographystyle{ieee_fullname}
\bibliography{iclr2021_conference}
}

\begin{appendix}
\onecolumn
\section{Models parameters for each experiment}

\begin{table}
\centering
\begin{small}
\begin{tabular}{lclclclcl} 
\toprule
    Experiment short name & & One Hot & & StackMix & & \% Difference \\ 
    \cmidrule{1-1} \cmidrule{3-3} \cmidrule{5-5} \cmidrule{7-7}
    \texttt{RN50-IMAGENET} & & $25,557,032$ & & $25,557,032$ & & $0\%$ \\
    \texttt{RN56-TINYIMAGENET} & & $1,865,768$ & & $2,070,568$ & & $10.9\%$ \\
    \texttt{VGG16-CIFAR100}  & & $15,038,116$ & & $15,300,260$ & & $1.7\%$ \\
    \texttt{PRN18-CIFAR100/-AA/-INF} & & $11,222,244$ & & $11,222,244$ & & $0.0\%$ \\
    \texttt{SRN18-CIFAR10} & & $11,267,842$ & & $11,267,842$ & & $0.0\%$ \\
    \texttt{RN20-CIFAR10} & & $570,602$ & & $573,162$ & & $0.4\%$ \\
    \texttt{WRN-STL10} & & $11,002,330$ & & $11,048,410$ & & $0.4\%$ \\
    \texttt{WRN-CIFAR10/100-C} & & $2,255,156$ & & $2,267,956$ & & $0.5\%$ \\
    \texttt{WRN-CIFAR10-SSL} & & $1,467,610$ & & $1,467,610$ & & $0.0\%$ \\
    \texttt{RN20-CIFAR10-N} & & $570,602$ & & ($k=2$) $573,162$ & & $0.4\%$ \\
    & & & & ($k=3$) $575,722$ & & $0.9\%$ \\
    & & & & ($k=5$) $580,842$ & & $1.8\%$ \\
    \texttt{VGG16-CIFAR100-N} & & $15,038,116$ & &  ($k=2$) $15,300,260$ & & $1.7\%$ \\
    & & & & ($k=3$) $15,562,404$ & & $3.4\%$ \\
    & & & & ($k=5$) $16,086,692$ & & $6.9\%$ \\
 \bottomrule
\end{tabular}\vspace{-0.1cm}
\caption{ Model Parameters for each experiment.}\vspace{-0.1cm}
\label{tableModelParams}
\end{small}
\end{table}
\end{appendix}

\end{document}